# Progressive enhancement and restoration for mural images under low-light and defected conditions based on multi-receptive field strategy

Xiameng Wei[a]†, Binbin Fan[b]†, Ying Wang[b], Yanxiang Feng[b]* and Laiyi Fu[b,c,d]*

[a] School of Design and Art, Xijing University, Xi'an,Shaanxi 710000, China
[b]Systems Engineering Institute, School of Electronic and Information Engineering, Xi'an Jiaotong University, Xi'an,Shaanxi 710049, China
[c]Research Institute of Xi'an Jiaotong University, Zhejiang, 311200 Hangzhou, China
[d]Sichuan Digital Economy Industry Development Research Institute, 610036 Chengdu, China
* Corresponding author
†The authors want to claim that the first two authors should be considered as equal contribution.
laiyifu@xjtu.edu.cn

**Abstract.** Ancient murals are valuable cultural heritage with great archaeological value. They provide insights into ancient religions, ceremonies, folklore, among other things through their content. However, due to long-term oxidation and inadequate protection, ancient murals have suffered continuous damage, including peeling and mold etc. Additionally, since ancient murals were typically painted indoors, the light intensity in images captured by digital devices is often low. The poor visibility hampers the further restoration of damaged areas. To address the escalating damage to ancient murals and facilitate batch restoration at archaeological sites, we propose a two-stage restoration model with automatic defect area detection strategy which called MER(Mural Enhancement and Restoration net) for ancient murals that are damaged and have been captured in low light. Our two-stage model not only enhances the visual quality of restored images but also achieves commendable results in relevant metric evaluations compared with other competitors. Furthermore, we have launched a website dedicated to the restoration of ancient mural paintings, utilizing the proposed model. Code is available at https://gitee.com/bbfan2024/MER.git.

## 1   INTRODUCTION

Murals, created on walls through drawing, sculpture, or other painting methods, represent one of the earliest forms of human art. Over the centuries, Chinese mural paintings have developed into a vast collection, holding immense artistic, cultural, and historical value. These murals are a treasured embodiment of Chinese civilization.

The structural and textural information of murals inspire modern painting techniques and artistic styles which scholars are capable of analyzing to understand ancient societies' aesthetics, life, religion, architecture, and folklore [1]. However, natural aging, biological aggression, man-made pollution have led to cracks, partial detachment, and mold growth in many ancient murals, causing ongoing damage. But the restoration of handmade murals requires a great deal of specialized knowledge and is slow. In addition, hand-painting on original murals is irreversible [2].

Recently, the conservation ancient murals using computer technology has garnered significant attention among scholars, and digital restoration methods have proven to be both more efficient and reversible[3]. Therefore, applying more efficient deep learning-based digital image restoration techniques to mural paintings' restoration is critically urgent, aiming to repair the damage to mural paintings as thoroughly as possible without destruction and to record the information they contain [4]. Unlike traditional image restoration problems, the restoration of ancient murals has the following challenges: (1) Murals collection often occurs in dimly lit environments, such as ancient mural sites like tombs and temples. Camera flashes or continuous lighting will cause irreversible damage which are prohibited near murals to preserve them. Dark environments result in low-light images with poor brightness and pixel values which existing image restoration networks struggle with providing inadequate results. The challenge lies in enhancing the network's ability to extract features from low-light images, repair damage, and create an image that retains its historical and cultural significance while being more visually accessible. (2) Characteristics of mural data increase training difficulty. Due to murals' unique colors and textures, directly adapting pre-trained models which applied in other scenes to murals is ineffective. Training a mural-specific network requires extensive labeled data, which is scarce. Moreover, murals exhibit various defect types with random distribution which traditional restoration networks struggle to adapt to. (3) Traditional image restoration networks typically handle specific image sizes and struggle with large mural images. But murals come in a wide variety of colors and huge sizes. So many studies have made attempts to apply traditional image restoration techniques to murals' restoration but there is no effective deep learning network in the field of murals' restoration that can balance low-light enhancement and mural inpainting for giant ancient murals.

We divided the task into two parts: murals enhancement and murals inpainting. In our research, we found no existing work that effectively combines these two components specifically for mural restoration. Therefore, we will review generalized visual image enhancement and inpainting methods separately. Earlier image enhancement was achieved by adjusting the brightness and contrast of an image based on histogram equalization [23]. Although the method consumes less computational resources, histogram equalization is sensitive to noise in the image and does not work well in dealing with more complex image situations, such as problems in regions with large local contrast differences. And deep learning models can solve the deficiencies that occur in histogram equalization methods. Deep learning models can be trained based on a large amount of data, have better generalization ability and adaptability, and can object local features of an image to better handle image enhancement tasks of different types and complexity. Lore et al.[24] pioneered the use of deep architectures for low-light image enhancement. Zhang et al.[25] inspired by Retinex theory constructed a deep neural network to reconstruct reflectance and illumination maps. Wang et al.[26] introduced intermediate illumination and proposed a new end-to-end network. Yang et al.[27] proposed a deep recursive band network applying adversarial learning to recover low-light images. Zhang et al.[28] proposed a maximum entropy based Retinex model to estimate illumination and reflectance.

The current image inpainting methods are roughly categorized into two kinds: traditional image restoration methods and deep learning-based image restoration methods. Early image restoration often relied on diffusion techniques. Studies [5, 6, 7, 8] reconstructed missing areas by diffusing semantic information from surroundings domains into the gaps. Bertalmio et al.[5] were the first to apply anisotropic diffusion iteratively for image restoration, enhancing clarity. However, the diffusion method focusing only on information near the missing area, struggles with large gaps. In contrast, patch-based methods are able to recover larger regions. Patch-based methods fill gaps by transferring similar parts from neighboring images. A. A. Efros and T. K. Leung [9] pioneered the patch-based method, synthesizing textures in gaps by reusing local textures from the input image. In order to reduce the time cost of patch matching, Barnes et al.[6] proposed a stochastic nearest-neighbor patch matching algorithm based on the patch method. However, patch methods depend on surrounding content to inpaint damaged areas. If the surrounding texture lacks similarity or is too complex, desired inpainting results are unattainable.

With the rapid development of deep learning techniques, image restoration methods have become a popular and effective option. These methods can be broadly categorized into the following approaches: progressive, structural information-guided, attention-based, convolution-based, and diversified restoration. Progressive restoration approaches are step-by-step, gradually restoring the missing regions. Yu et al.[16] used a coarse-to-fine strategy, where a simple dilated convolutional network is used to make a coarse prediction, and then a second network takes the coarse prediction as input and predicts a finer result. Zhu et al.[21] proposed a coarse-to-fine cascade refinement architecture. Li et al.[17] utilized a part-to-full strategy, where the simpler part is solved first parts, and then use the results as additional information to progressively enhance the constraints to solve the difficult parts. Structural information-guided interior mapping relies on the structure of known regions and Yang et al.[18] used a gradient-guided restoration strategy. Attention mechanism based image restoration strategies utilize the concept of human attention to improve the accuracy, efficiency and retention of semantic information in restoration algorithms.Yu et al.[16] used contextual attention based approach. Zeng et al.[19] used attention transfer based strategy and proposed a pyramid contextual encoder, which progressively learns high-level semantic feature maps by paying attention to the region affinity and transfers the learned attention to its neighboring high-resolution feature maps.inpaint with convolutional perception uses masks to indicate missing regions, Xie et al.[20] used a bidirectional convolution-based strategy and proposed a learnable attention graph module for efficiently adapting to irregular nulls and propagation in convolutional layers. Diversified restoration methods are to generate multiple visually plausible results. Zheng et al.[22] proposed a novel framework with two parallel paths that can generate multiple different solutions with reasonable content for a single mask input. However, most of the above image restoration methods are aimed at portraits, street scenes, etc., and cannot be directly used in the field of mural restoration due to the special characteristics of the application scene of murals restoration.

In the field of murals restoration, Cao et al.[40] used Generative Adversarial Networks to repair missing mural regions and used the introduction of global and local discriminative networks to evaluate the restored images. Lv et al.[41] separated the contour line pixel regions and content pixel regions of a mural image by connecting two generators based on U-Net. Merizzi et al.[42] used Deep Image Prior method to repair damaged murals; Li et al.[39] proposed line-drawing guided progressive murals restoration method; Song et al.[43] proposed AGAN, an automatic coding generative adversarial network for image restoration of Dunhuang wall paintings. The above murals restoration methods perform well in structural reconstruction and color correction under conventional conditions, but none of them take into account the actual low-light environment of the defective murals.

In this paper, we aim to address the current two main challenges (low light condition as well as damaged condition) simultaneously, utilizing a two-stage framework for batch restoration of ancient murals in archaeological settings. We propose MER(Mural Enhance and Restoration) network, designed to restore defective murals captured under low-light conditions, yielding high-quality imaging restoration results. Our approach involves a two-stage processing strategy, particularly suited for low-light mural images characterized by low pixel values and difficulties in texture and structural feature extraction, hampering contextual learning and restoration by neural networks.

Our neural network comprises two main stages: low-light image enhancement stage and restoration stage. The initial stage employs weight-sharing cascaded illumination learning and a self-calibration module to enhance low-light images, improving clarity and aiding structural and textural recognition. This enhancement enables the second-stage restoration network to extract image texture and structural features effectively. In the second stage, the brightness-enhanced image is processed by an image restoration network with varying receptive fields, catering to both global and local defect inpainting. Incorporating discriminators and diverse loss functions produces more realistic restorations, enhancing overall image quality for improved human perception of texture and structure.

Finally, the contributions of this article are as follows:
- We propose the MER, a two-stage murals restoration model that effectively handles defect inpainting while considering the impact of low-light imaging, resulting in higher-quality restoration of ancient murals.
- We apply low-light processing and segmentation strategies for effective dataset expansion, where the segmentation strategy enhances both training and inpainting outcomes.

- We design the gradient based strategies for multiple image defect areas detection, to identify areas requiring restoration for more real-world scenarios, which enhances the applicability of the model.
- We develop a free available web server with pre-trained MER model as backend for end users to restore mural images online.

## 2 METHODOLOGE

### 2.1 Flaw Finding

In practical applications it is necessary to analyze the damaged ancient wall paintings to find the flaw region. In order to find the flaw region, during the inpainting network, we design a simple yet effective strategy to describe the defective region.

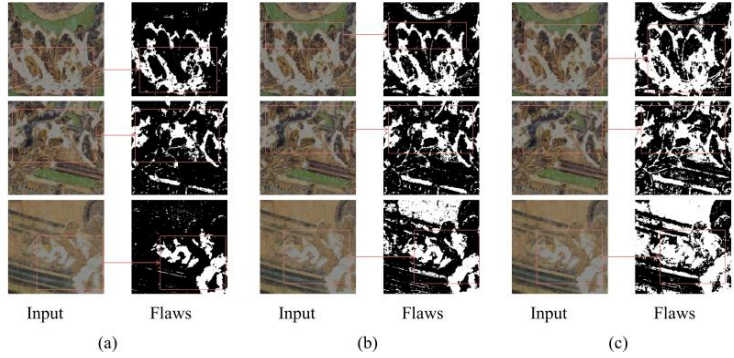

**Fig. 1.** Example of Masking Generation for Defective Images in a Real Application Scenario. (a)Damage markers obtained using the image gradient method when $\lambda_G$=4, $\lambda_P$=3. (b)Damage markers obtained using the image gradient method when $\lambda_G$=4, $\lambda_P$=2. (c)Damage markers obtained using the image gradient method when $\lambda_G$=5, $\lambda_P$=1.5.

First, we calculate the pixel gradient for the ancient mural images. We reason that if the color of the defective region does not belong to the original image, then the boundary of the defective region can be obtained by screening the outliers of the pixel gradient. The threshold $G_{th}$ of the gradient outlier is defined as follows:

$$G_{th} = G_{avg} + \lambda_G \sigma_G \tag{1}$$

Where $G_{avg}$ is the mean of the pixel gradient, $\sigma_G$ is the standard deviation of the pixel gradient and $\lambda_G$ is the threshold parameter of the gradient, which was set to 3 in the experiment.

After getting the boundary of the defective region we calculate the pixel values of the boundary. We consider the pixel values of the defects as similar and outliers, so the threshold $P_{th}$ for pixel outliers is defined as follows:

$$P_{th} = P_{avg} + \lambda_P \sigma_P \tag{2}$$

Where $P_{avg}$ is the mean value of the pixel value at the edge of the defect, $\sigma_P$ is the standard deviation of the pixel value at the edge of the defect, and $\lambda_P$ is the threshold parameter of the pixel value of the defective region was set to 2 in the experiment.

Finally, we calculate the ancient mural images in channels and defined the regions with pixel values higher than the pixel value threshold $P_{th}$ as defective regions, from which we generated to get the defect description map.

### 2.2 Low-Light Image Enhancement Stage

Given that murals are typically found in indoor areas such as grottoes, tombs, temples, etc., we notice that the sites of real murals conservation tend to have low brightness, resulting in captured mural images being in low-light conditions. Low-light images exhibit poor visibility, and even after damaged areas are restored, the most valuable texture information in the murals remains difficult to discern. To recover valuable mural images, we design a low-light image enhancement network. In classical Retinex theory $y = z \otimes x$, we know that there is a link between the low light observation $y$ and the desired clear image $z$, where $x$ denotes the illumination component and it is the core optimization objective in low-light image enhancement operations. According to Retinex theory, better results can be obtained by getting a good estimate of the illumination. Given that illumination and low-light observations in murals are often similar or linearly related, and considering the need to minimize model parameters due to dataset limitations, we are inspired by the work [29] to use a light estimation network $\mathcal{H}_\theta$ with shared structure and parameters to form the augmentation module $E$, where $\theta$ is a learnable parameter. To ensure outputs from different rounds of progressive optimization converge to the same state, we adopt a network $K_\vartheta$ with learnable parameters $\vartheta$ to form a self-calibrating module $S$ progressively corrects the inputs of the augmentation module in each round. The formulation of the progressive optimization network is expressed as follows:

$$E(v^t): \begin{cases} u^t = \mathcal{H}_\theta(v^t), v^0 = y \\ x^{t+1} = v^t + u^t, \end{cases} \quad (3)$$

$$S(x^k): \begin{cases} z^k = y \oslash x^k, \\ v^k = y + K_\vartheta(z^k), \end{cases} \quad (4)$$

Where $k = t + 1$, $v^t$ is the transformed input in round $t$ of the asymptotic process, $u^t$ is the residual in round $t$, and $x^t$ is the illumination in round $t$, ($t$=0,1...).

Furthermore, in order to better train the low-light image enhancement network, we refer to the work [30,31] which use an unsupervised loss function $\mathcal{L}_E$ defined as:

$$\mathcal{L}_E = \alpha \sum_{t=1}^{T} ||x^t - (y + s^{t-1})||^2 + \beta \sum_{i=1}^{N} \sum_{j \in N(i)} \omega_{i,j} |x_i^t - x_j^t|, \quad (5)$$

$$\omega_{i,j} = exp\left(-\frac{\sum_c \left((y_{i,c}+s_{i,c}^{t-1})-(y_{j,c}+s_{j,c}^{t-1})\right)^2}{2\sigma^2}\right), \quad (6)$$

Where $T$ is the total number of optimization enhancement rounds, which we set to 6 in our experiments. $N$ is the total number of pixels, $s^{t-1}$ is the output of $K_\vartheta$, $i$ is the $i-th$ pixel, $c$ is the image channel in the *YUV* color space and σ = 0.1 is the standard deviation of the Gaussian kernel. α and β are the weights which we set to 1.5 and 1 in our experiments.

## 2.3 Image Inpainting Stage

The final output from the low-light image enhancement stage serves as the input $I_{in}$ for the image restoration network, tasked with restoring damaged areas. Previous work has shown that networks with larger receptive field better restore overall structure and larger color blocks, while those with a smaller receptive field excel at restoring detailed local textures. Inspired by the work [32], we employ networks with varying receptive fields for distinct restoration tasks, following a coarse-to-fine restoration strategy. The receptive field is the set of input pixels which are path-connected to a neuron.

In order to roughly recover the structure of the broken mural as a whole, we first input the mural $I_{in}$ which is to be restored with a binary mask $M$ describing the missing regions (where 0 denotes a valid pixel and 1 denotes a missing pixel) into a coarse inpainting network ($Net_C$) with a large receptive field. Where the mask $M$ is randomly generated in training process and automated by our flaw finding method in real application scenarios. $Net_C$ includes eight downsampling and upsampling operations, utilizing long skip connections to transfer information between the encoder and decoder, thereby recovering information lost during downsampling. The output of $Net_C$ is $I_{out}^C$. Additionally, we incorporate a patch-based discriminator with spectral normalization [33] to improve the realism of the restoration. This discriminator accepts the real mural and the restored mural as inputs and produces a two-dimensional feature map of size $\mathbb{R}^{32\times 32}$ as output. In this feature map, each element distinguishes between real and artificially generated content. The formulae for the spliced image and pixel-level loss of the rough restoration process are expressed as follows:

$$I_{mer}^C = I_{in} \odot M_r + I_{out}^C \odot M, \tag{7}$$

$$\mathcal{L}_r^C = \frac{1}{sum(M_r)}||(I_{out}^C - I_{gt})\odot M_r||_1 + 6\frac{1}{sum(M)}||(I_{out}^C - I_{gt})\odot M||_1, \tag{8}$$

Further, we can define the relevant loss in the GAN method as follows:

$$\mathcal{L}_G^C = 0.1\mathbb{E}_{I_{mer}\sim PI_{mer}(I_{mer})}[(D(I_{mer}^C) - 1)^2], \tag{9}$$

$$\mathcal{L}_D = \frac{1}{2}\mathbb{E}_{I\sim Pdata(I)}[(D(I_{gt}) - 1)^2] + \frac{1}{2}\mathbb{E}_{I_{mer}\sim PI_{mer}(I_{mer})}[(D(I_{mer}^C))^2], \tag{10}$$

Where $M_r$ is the result of inverting the $M$ pixel value (i.e., 0 becomes 1 and 1 becomes 0), $I_{mer}^C$ is the merged image, $I_{out}^C$ is the output of $Net_C$, $I_{gt}$ is the ground-truth mural corresponding to the input, $\mathcal{L}_r^C$ represents the pixel-wise reconstruction loss of the coarse-inpainting network, $\mathcal{L}_r^C$ represents the adversarial generation loss of the coarse-inpainting network, and $\mathcal{L}_D$ represents the loss of the discriminator.

After completing the coarse restoration, we need to inpaint the local structure and texture of the mural in further detail. We input the coarsely restored mural into a local restoration network $Net_L$ with a small receptive field . This network consists of two up-sampling operations, four residual blocks and two down-sampling operations to process the local regions of the input image in a sliding window manner, which can better inpaint the local information of the mural. The pixel-wise reconstruction loss of the localized restoration step is $\mathcal{L}_r^L$, which is the same process as the $\mathcal{L}_r^C$ operation, except that $I_{out}^C$ is replaced by the output $I_{out}^L$ of $Net_L$ in Eqn.(8). $I_{mer}^L$ is also merged in the same way as $I_{mer}^C$ in the coarse restoration step, except that $I_{out}^C$ is replaced by the output $I_{out}^L$ of $Net_L$ in Eqn.(7). And we use the attention module in this process. Further, we are inspired by the work [34,38] to use the total variation loss ($\mathcal{L}_{tv}^L$) for smoothing neighboring pixels in the hole region, and the perceptual loss ($\mathcal{L}_{per}^L$) [36] and style loss ($\mathcal{L}_{sty}^L$) [37], which are computed in feature space as defined on VGG-16 [35]. Their combination with $\mathcal{L}_r^L$ to form the object loss $\mathcal{L}_L$ will achieve better recovery. The added pixel-level loss function can be defined as follows:

$$\mathcal{L}_{tv}^L = ||I_{mer}^L(i,j+1) - I_{mer}^L(i,j)||_1 + ||I_{mer}^L(i+1,j) - I_{mer}^L(i,j)||_1, \qquad (11)$$

The loss function based on feature space computation can be defined as follows:

$$\mathcal{L}_{per}^L = \sum_i ||\mathcal{F}_i(I_{out}^L) - \mathcal{F}_i(I_{gt})||_1 + ||\mathcal{F}_i(I_{mer}^L) - \mathcal{F}_i(I_{gt})||_1, \qquad (12)$$

$$\mathcal{L}_{sty}^L = \sum_i ||\mathcal{G}_i(I_{out}^L) - \mathcal{G}_i(I_{gt})||_1 + ||\mathcal{G}_i(I_{mer}^L) - \mathcal{G}_i(I_{gt})||_1, \qquad (13)$$

Where $\mathcal{F}_i$ represents the i-th layer feature map of the pre-trained VGG-16 network, $\mathcal{G}_i = \mathcal{F}_i(\cdot)\mathcal{F}_i(\cdot)^T$ is the Gram matrix [39]($i \in \{5,10,17\}$).

$$\mathcal{L}_L = \mathcal{L}_r^L + \lambda_{tv} \cdot \mathcal{L}_{tv}^L + \lambda_{per} \cdot \mathcal{L}_{per}^L + \lambda_{sty} \cdot \mathcal{L}_{sty}^L \qquad (14)$$

Finally, the total loss function of this repair step can be defined as Eqn.(14), where $\lambda_{tv}$, $\lambda_{per}$ and $\lambda_{sty}$ are the weight parameters which were set to 0.1, 0.05 and 120 in the experiments respectively.

After completing the local texture and structure refinement inpainting, the mural presents a better texture inpainting effect. In order to get a better overall visual effect, we input the output mural of the local refinement network to the global refinement network $Net_G$. $Net_G$ adds three attentional modules in front of the decoder of $Net_C$. The attentional mechanism leads to a better effect. The training loss function $\mathcal{L}_G$ of $Net_G$ is similarly to that of $\mathcal{L}_L$ by simply replacing $I_{out}^L$ with $I_{out}^G$. Finally, we can get the training loss $\mathcal{L}_R$ of the whole image restoration network as the sum of the three restoration subnetworks' losses and the sum of the GAN work losses, i.e., $\mathcal{L}_r^C + \mathcal{L}_G^C + \mathcal{L}_D + \mathcal{L}_L + \mathcal{L}_G$.

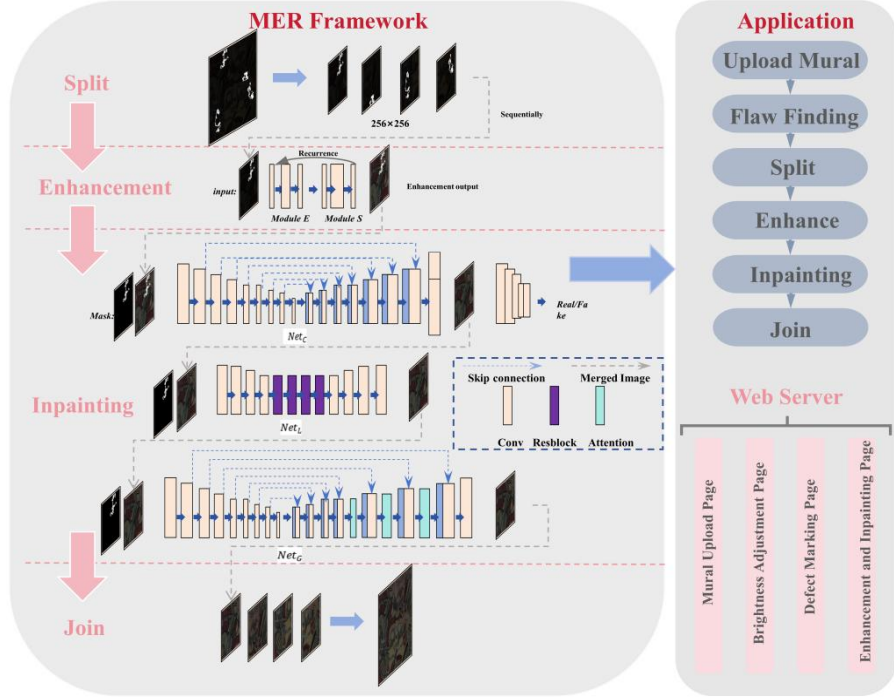

**Fig. 2.** The architecture of the MER. The mural to be restored is first divided into small images of 256×256 pixel size and sequentially fed into the subsequent restoration model. The corresponding mask describing the defect will then be automatically generated. Murals that are displayed in low light and contain defects will be light-enhanced first, and then the output of the image enhancement network will be fed into the inpainting network along with the mask for further restoration, as specified in the image merging equation in Eqn.(7). After completing the enhancement and inpainting of the small images, we will join the small images to get the final result. The violet block within the local refinement network ($Net_L$) signifies a residual block consisting of two layers. Meanwhile, the global refinement network ($Net_G$) incorporates three green blocks, each representing attention blocks with resolutions of 16×16, 32×32, and 64×64 respectively.

### 2.4 Parameter Updating Strategy

We then propose a two-stage strategy to seamlessly achieve the two desired outcomes: restoring ancient murals damaged under low-light conditions into images with enhanced observability. The output of the low-light image enhancement phase serves as the input for the image restoration phase, creating a coupling between the two phases. During training we employ an optimizer to adjust network parameters based on the loss, aiming to minimize the difference between the generated output and the target. However, if both phases update parameters simultaneously, the input to the image restoration phase becomes unstable due to ongoing changes in the enhancement phase's network parameters. This increases the learning difficulty in the image restoration stage. At the same time, we recognize the differing complexities of the network's two phases and the varying number of learning samples required for optimization.

Thus we decide to use an alternating parameter update strategy. In one epoch we divide the dataset into several subsets, each with 60 data points. The first 6 data points in each subset are dedicated to optimizing the enhancement network, during which we freeze the image restoration network's parameters to focus solely on updating the low-light image enhancement network. The remaining 54 data in the data set are used for the restoration network optimization phase, where we optimize the image inpainting network's parameters while freezing those of the low-light image enhancement network.

Further, the loss function of the whole MER $\mathcal{L}_{MER}$ can be expressed as follows:

$$\mathcal{L}_{MER} = \lambda_R \mathcal{L}_R + \lambda_E \mathcal{L}_E \tag{15}$$

Where parameter $\lambda_R$ will set to 1 in the inpainting network optimization phase and 0 in the enhancement network optimization phase, and parameter $\lambda_E$ will set to 0 in the inpainting network optimization phase and 1 in the enhancement network optimization phase.

## 3 Experimental

This section begins with an overview of the mural restoration outcomes achieved using the MER model. Subsequently, we detail the specific experiments, covering the dataset, experimental setup, and evaluation metrics. The visual outcomes and metric data from these experiments demonstrate the MER model's effective murals restoration capabilities.

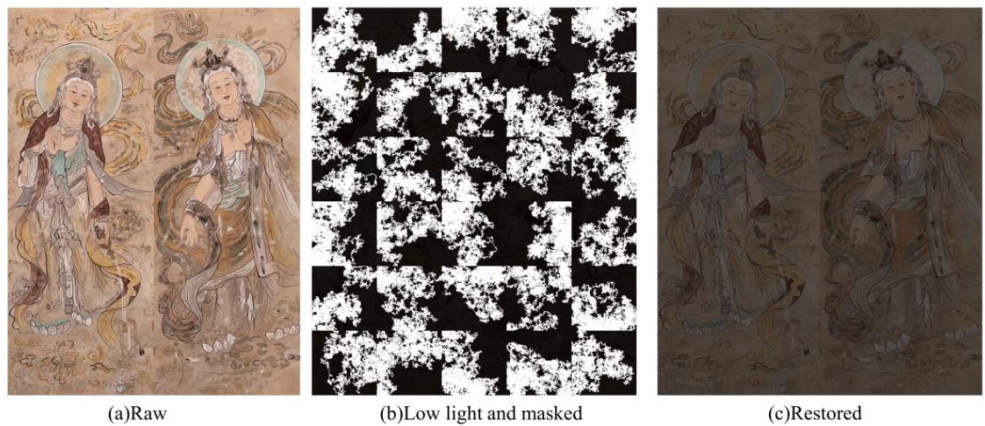

**Fig. 3.** (a) Raw refers to the original image of an ancient mural painting. (b)Low light and masked refers to the low-light and damaged mural image obtained by reducing the brightness of the original image to 12% and adding random masking to simulate the real archaeological scene. (c) Restored refers to the image after being restored by the MER network.

## 3.1 Experimental settings

**Dataset.** In order to achieve the best possible murals restoration results, we collect data on different styles of murals and expand the dataset.

*Mural data.* We first gathered captured images of famous mural paintings from China, which include images from Guangsheng Temple in Hongdong, water and land mural paintings at Princess Temple in Fanchi, and mural paintings at Yongle Palace in Ruicheng. Due to murals typically being located on indoor walls of tombs, grottoes, temples, and palaces, and restrictions on using excessive light to protect them, many ancient murals are scanned and preserved under darker imaging conditions, making the image data hard to observe. To mimic the lower brightness observed in real-world data collection scenarios, we artificially reduced the brightness of the collected images to 55%, 37%, and 12% of the brightness of the original images. Additionally, to enrich the dataset while considering the final restoration strategy, we crop the three obtained datasets with different low brightness into 256*256 pixel sub-images. The final obtained low-light subimages for training are 123,299, and low-light subimages for testing are 4,581.

*Mask data.* Research into the damage causes of ancient murals across various regions revealed common factors: stains, corrosion, peeling, mold, and cracks. To closely mimic real-world scenarios in experiments, based on identified damage types, we employed a random walk method to generate Dusk masks with random positions and shapes, simulating corrosion damage. Similarly, we used a corrosion operation to create Jelly masks, simulating edge damage around Dusk masks [37]. To simulate peeling damage, Droplet masks are created by randomly generating scattering points, simulating dispersed mold spots and liquid splashes. Block masks, generated from large circular blocks, mimic mural stains, while Line masks, created through random straight lines, also simulate stain damage. The cracks and scratches of the ancient murals are simulated in the real application scenario. To replicate the diversity of damage in real scenarios, masks covering 5% to 50% of the area were generated. Various mask shapes and sizes were integrated into the network for both training and testing.

**Experimental environment.** The GPU used for the experiments is NVIDIA's TITAN RTX. The Adam optimizer with a batch size of 6 is used to train our network. We set the initial value of the learning rate to 0.0001 for the first 100 trainings and linearly decayed it to zero for the next 100 trainings.

**Metics.** In our experiment, we use three mostly used evaluation metrics to evaluate the performance of the restoration model by calculating the original image and the low-luminance image processed by the restoration model. These metrics include: Peak Signal-to-Noise Ratio (PSNR): to evaluate image distortion; Structural Similarity (SSIM): to give a combined image similarity score by comparing factors such as luminance, contrast, and structural attributes as well as Learned Perceptual Image Patch Similarity (LPIPS): a deep-learning based image quality evaluation metric that simulates visual similarity in human perception.

### 3.2 Comparison with state-of-the-art relevant methodologies

In our research, we find that there is a lack of effective open source projects that can accomplish both light enhancement and defect repair in the field of murals' restoration. Therefore, we focus more on the imaging effect of defect restoration. We select seven representative state-of-the-art correlation methods to compare with ours. We perform experiments using images of three light intensities with three different masking area ranges and made quantitative and qualitative comparisons: the MADF[21], LG[32], PEN[19], RFR[16], MuralNet[39], AGAN[43] and SCI[29], where MuralNet[39] and AGAN[43] are recent works particularly designed for murals restoration.

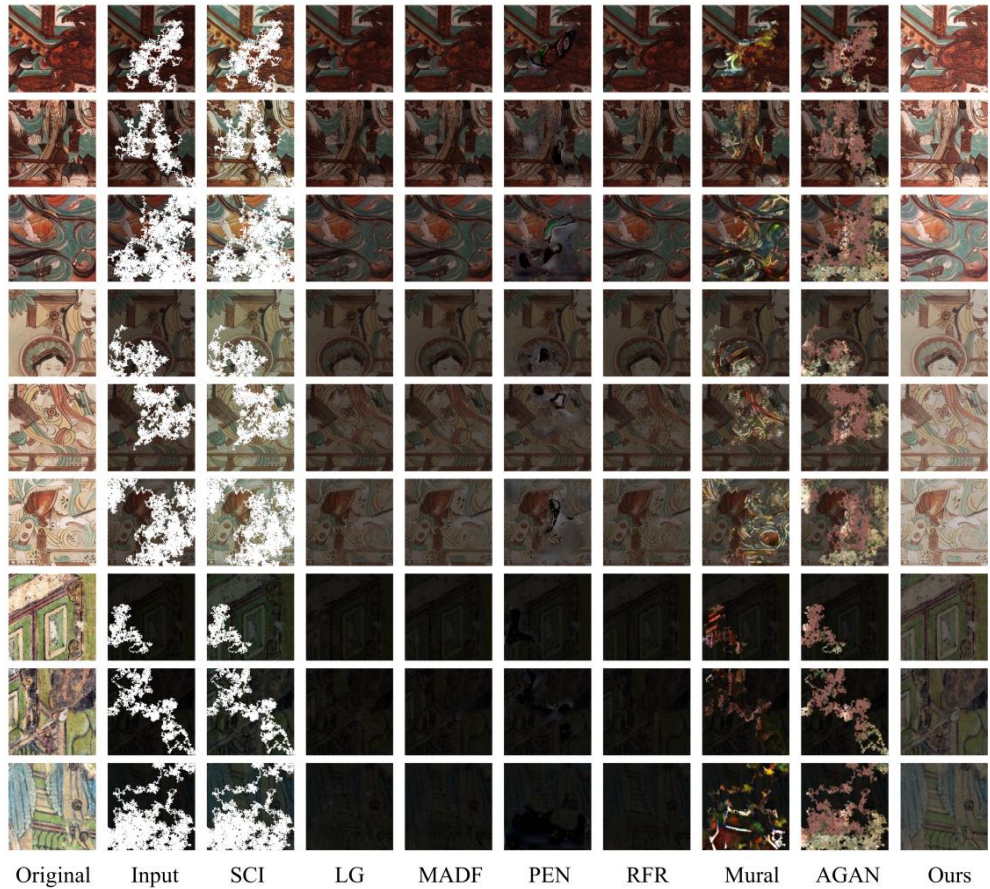

**Fig. 4.** Comparison of our method with SCI[29], LG[32], MADF[21], PEN[19], RFR[16], MuralNet[39] and AGAN[43] at three different low luminosities and applying three different areas of masking.

Qualitative comparisons is shown in Fig.4, which presents the visual outcomes of our method alongside SCI[29], LG[32], MADF[21], PEN[19], and RFR[16], processed under three different low-light conditions and with three varying areas of masking. The SCI[29] network merely brightens low-light images without restoring them, indicating it lacks image restoration capabilities. This implies the SCI[29] network alone cannot fully recover the content of damaged images in low light. Moreover, the color of the images processed with SCI[29]'s pre-trained model significantly deviates from the ground truth's color, highlighting the murals' unique color style. Additionally, we compare and analyze the image restoration results of other models. PEN[19] struggles with larger masked areas, showing holes and significantly worse restoration effects, indicating it's not suited for extensive damage restoration. MADF[21] outperforms PEN[19] in overall restoration but falls short in accurately inpainting textures at localized defect areas. Similarly, RFR[16] exhibits poor restoration in areas with significant pixel gradient variations in murals. LG[32] achieves a better overall imaging effect, yet its brightness and colors diverge considerably from the ground truth. Despite LG[32], MADF[21], PEN[19], RFR[16] in the damaged areas, their restored images at low brightness levels fail to reveal detailed textures and structures, especially evident when the ground truth's brightness is reduced to 12% in experiments. Consequently, these restoration networks cannot be directly applied to real-world ancient murals restoration scenarios. And, we can find that the line drawing guild method of MuralNet [39] cannot properly assist restoration in low light conditions, even though it is a restoration network for murals. Similarly, AGAN's [43] restoration effects are not balanced and do not respond effectively to low-light conditions.

In contrast, our proposed MER effectively restores damaged regions, improves both image brightness via the enhancement network, and ensures textures and structures are discernible to the human eye. At the first two brightness levels, our restoration results closely resemble the ground truth, whereas the outcomes from other networks appear blurred. At the lowest brightness level, the content in images restored by other networks becomes indiscernible, while MER's restoration allows clear observation of image content.

Quantitative evaluation also shows our experiment result, with three different ranges of random area masking for each of the three groups of low-light-intensity murals, with 1,527 test images in each group. We use three metrics, LPIPS(↓), PSNR(↑), and SSIM(↑), to evaluate the difference between the model processing results and the ground truth. The box-and-whisker plots for these metrics, based on the experimental results, allow us to clearly compare and conclude that our method's restored images surpass those of other related methods across all test datasets. Based on these findings, we can conclude that our proposed method achieves commendable results in restoring damaged murals in low-light conditions.

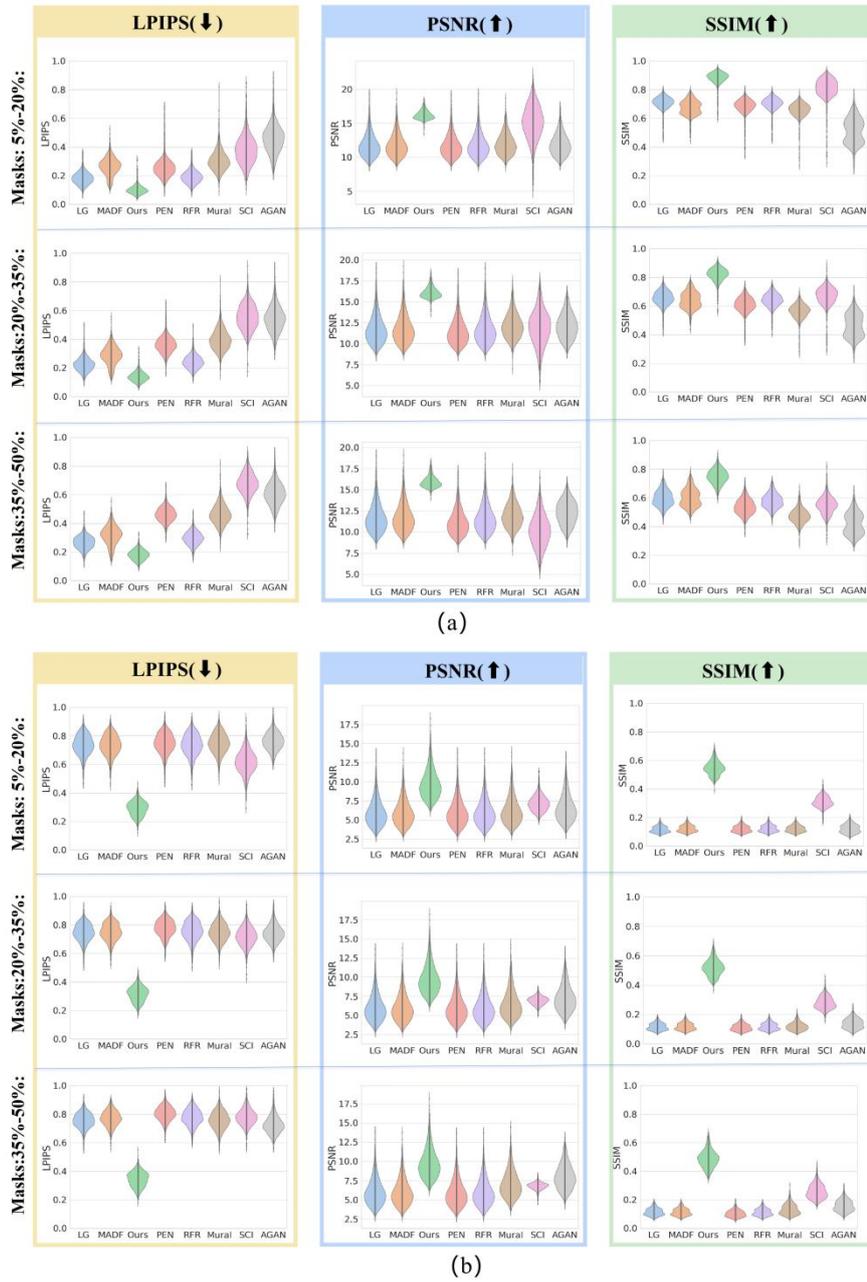

**Fig. 5.** Boxplots of the metrics evaluating the restorey results of our method with the other six correlation methods for different masking areas. Where (a) the input luminance is 55% of the ground truth and (b) the input luminance is 12% of the ground truth. For LPIPS metric, the lower is better. For PSNR and SSIM metrics, the higher is better.

### 3.3 Performance evaluation experiments on MER

**Enhancement Performance.** We calculate the average pixel values of the three low-brightness images with the average pixel values of the brightness-enhanced images obtained after the brightness enhancement stage of MER. In the results of TABLE 1 we can see that the MER significantly improves the brightness of the input image and achieves a sharper imaging effect.

**TABLE 1.** Average pixel values of the experimental results for the three low brightness cases

| Lightness | 55% | 37% | 12% |
|---|---|---|---|
| Low light | 78.7372205 | 52.59097298 | 16.33763162 |
| MER | 182.2823013 | 139.571474856 | 61.01057287 |

**Inpainting Performance.** We test the inpainting effect of our method under different types of masking. Fig.6 illustrates that MER effectively restores the structure and texture of murals under coverings simulating various realistic damages, demonstrating the model's robustness to different types of damage. This reinforces our confidence in using MER to inpaint ancient murals with various damage types in real-world scenarios. Additionally, comparing MER's restoration outcomes across different datasets shows that performance metrics remain stable across varying masking areas, indicating strong model robustness.

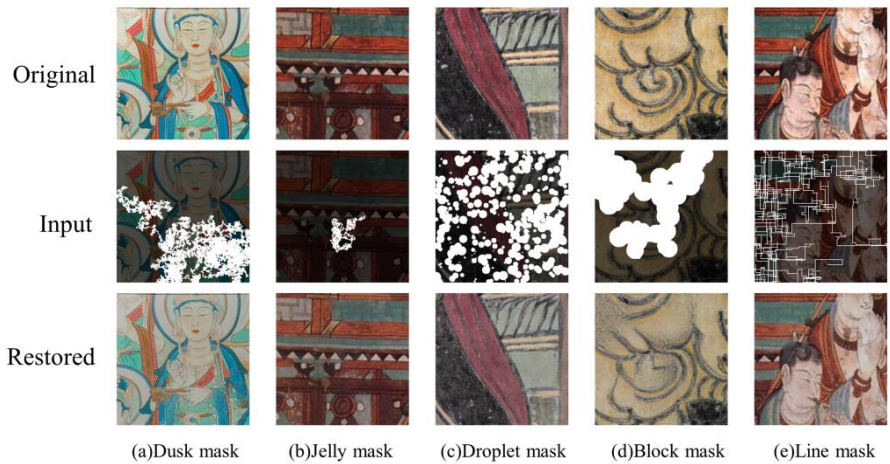

**Fig. 6.** Restore results of our method under different types of masks.

### 3.4 Ablation studies

To demonstrate the superiority of the MER network's two-stage strategy, we conduct ablation experiments comparing single-stage processing outcomes to those of two-stage processing. According to TABLE 3, using either the luminance enhancement or restoration stage alone fails to meet the requirements, with their performance metrics falling short of the two-stage processing strategy. Additionally, we analyze a 128x128 pixel segment of an experimental image, with its pixel value distribution displayed in Fig.7. The pixel distribution of the image from the two-stage strategy aligns more closely with the ground truth, as shown in Fig.7.

**TABLE 2.** Experimental results of model robustness under different defect coverage for three brightness cases

| Metrics | Lightness | Mask:5%-20% | Mask:20%-35% | Mask:35%-50% |
|---|---|---|---|---|
| LPIPS(↓) | 55% | 0.101882869 | 0.139504499 | 0.181546189 |
|  | 37% | 0.103287976 | 0.139249035 | 0.179793786 |
|  | 12% | 0.291675888 | 0.316039500 | 0.343960323 |
| PSNR(↑) | 55% | 16.25848322 | 16.12085126 | 15.94895506 |
|  | 37% | 24.02820589 | 23.17507427 | 22.33364033 |
|  | 12% | 9.746080353 | 9.764583779 | 9.777281565 |
| SSIM(↑) | 55% | 0.883252492 | 0.819436574 | 0.749673696 |
|  | 37% | 0.895213506 | 0.834300371 | 0.767272149 |
|  | 12% | 0.546299858 | 0.518490575 | 0.487885067 |

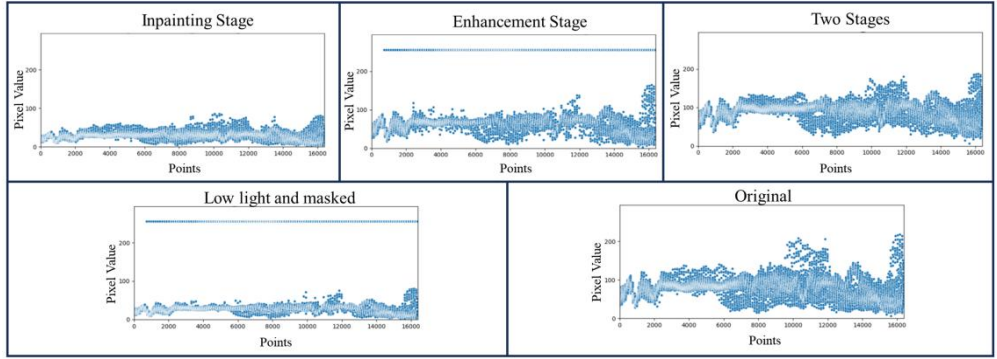

**Fig. 7.** Pixel distribution of an experimental image under different ablation experimental method.

### 3.5 Computational complexity of the MER

We select several representative networks to compare the processing times of the models for low-luminance and defective mural images as an indicator of model complexity. Firstly, we compare the processing speeds of various models within the inpainting framework on images with 37% brightness and masks covering 20%-35% of the area, with results presented in Fig.8. The processing speed of MER is notable for its commendability and stability, demonstrating good robustness. According to the data in TABLE 4, the model processes a single input image within 0.03s, enabling rapid restoration. Thus, the model's moderate computational complexity allows for efficient batch processing of scanned data at real archaeological sites.

**TABLE 3.** Ablation experiments at 20%-35% masking

| Metrics | Lightness | Enhancement Stage | Inpainting Stage | Two Stages |
|---|---|---|---|---|
| LPIPS(↓) | 55% | 0.555679194 | 0.223111976 | 0.139504499 |
|  | 37% | 0.590810317 | 0.331743732 | 0.139249035 |
|  | 12% | 0.720505025 | 0.747321599 | 0.3160395 |
| PSNR(↑) | 55% | 11.91393594 | 11.74744269 | 16.12085126 |
|  | 37% | 10.83945574 | 8.894324970 | 23.17507427 |
|  | 12% | 6.949973191 | 6.034000378 | 9.764583779 |
| SSIM(↑) | 55% | 0.670701967 | 0.656132778 | 0.819436574 |
|  | 37% | 0.638251357 | 0.447046324 | 0.834300371 |
|  | 12% | 0.280544982 | 0.117322734 | 0.518490575 |

**TABLE 4.** MER processing time for input murals (s)

| Lightness | Mask:5%-20% | Mask:20%-35% | Mask:35%-50% |
|---|---|---|---|
| 55% | 0.021101109 | 0.020688261 | 0.023290726 |
| 37% | 0.0287468 | 0.020363215 | 0.022794565 |
| 12% | 0.027878533 | 0.028666547 | 0.027223293 |

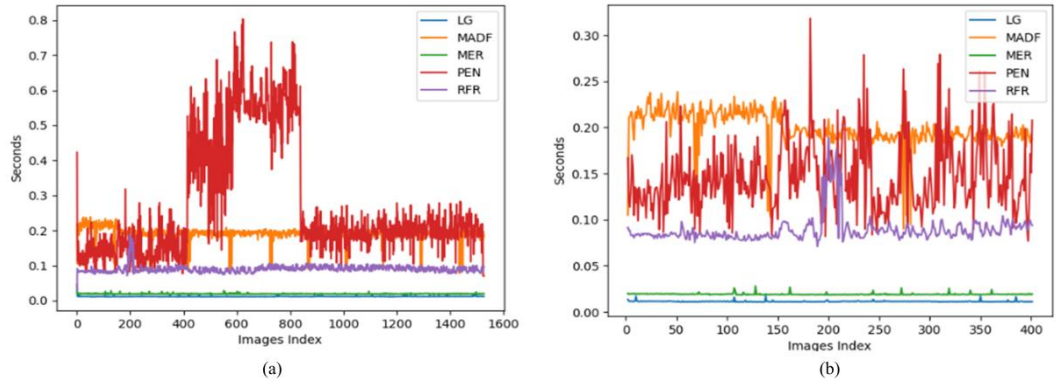

**Fig. 8.** Processing time of images containing masks of 20%-35% area size at 37% brightness for different inpanting networks. The figure shows the processing time for two randomized data batches, where the experimental batch in (a) is 1560 mural images and the experimental batch in (b) is 400 mural images.

## 3.6 Real Application Scenarios

In real ancient murals restoration scenarios, the acquired images of ancient murals are often large, low-light, and contain defects. Therefore, when applying MER to real ancient murals restoration, large images should first be split into 256x256 pixel sub-images for input into the restoration network. Additionally, it's necessary to identify defect information in the murals sub-images. We assume that defects share similar characteristics, with colors and shapes distinct from the original image content, resulting in significant gradient variation at the defect boundaries. In this experiment, a gradient operation a gradient operation is performed on the mural, using the mean and standard deviation of the gradient to identify outliers across the image, which are then considered as the boundaries of the damage. Additionally, we calculate the pixel values at the damage boundaries and use the mean and standard deviation to determine the range of pixel values for the damaged area, assuming that the damage will exhibit similar outlier pixel values. This process yields a damage marker map of the image. After inputting the mural sub-map and its damage marker map into the network for restoration, all sub-maps are reassembled to produce the restored ancient mural image. Of course, this step can be substituted with manual labeling to achieve similar restoration outcomes.

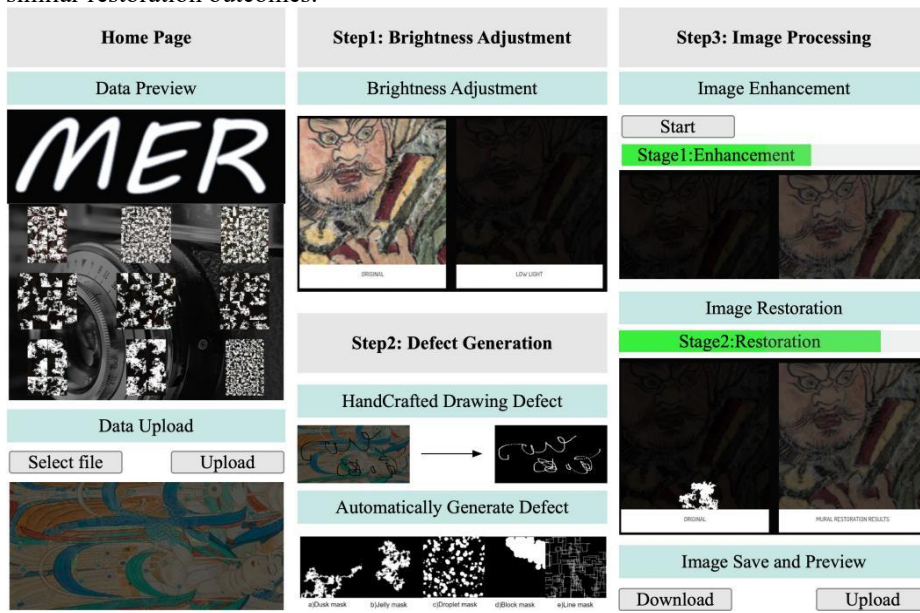

**Fig. 9.** MER web server interface. User can adjust the brightness of the mural which need to be restored on the Brightness Adjustment Page and can draw or add random defects on the Defect Generation Page. We also provide the option to automatically find defective regions to form a mask based on the gradient method. After completing the pre-operation, users can get the restored result of MER on the Image Processing Page and the results of the restoration will be presented in the order of the different stages.

To facilitate the accessibility of MER, we further develop a web server that runs MER on the backend using the Flask framework and made it freely available to users. Our server utilizes pre-trained MER models, allowing users to upload mural images for experimentation. We offer pages for low-light and defect simulation, enabling users to upload images of intact murals for simulation experiments. If the user's input image is already low-light and contains defects, the mural image can be repaired by defect labeling. The interface of our web server is shown in Fig.9.

## 4   Conclusions

In ancient murals conservation, ancient murals are often large, irregular and suffer from unpredictable damage. Scholars often use cameras to capture the valuable content of ancient murals for preservation. However, these images are often low-light and damaged, making it difficult to directly observe the murals' coherent texture and structure. To address this issue, we introduce the MER, a two-stage restoration network with the ability to automatically detect defected areas in murals. The first stage enhances the brightness of low-light mural images for improved visibility. The second stage employs a coarse-to-fine strategy to inpaint damaged areas, resulting in images with complete texture and structure. Moreover, in order to solve the limitation of large size and small dataset of ancient murals, we apply artificial low light processing and segmentation strategy. To improve the applicability of MER, we utilize various masks that simulate realistic application scenarios. A large number of experimental results have proved the commendable effectiveness of our proposed method.

However, our work is dedicated to propose a recovery method to solve the problem of low-light mural captured images with defective phenomena, and only a simple gradient calculation method is used in the application to obtain a mask image describing the defective regions. If improved recovery of damaged areas is needed, manual labeling is essential for accurately covering the defective regions.

**Acknowledgments.** This research was supported by Zhejiang Provincial Natural Science Foundation of China under Grant (No. LQ23F020018) (L.F), Natural Science Basic Research Program of Shaanxi (Program No. 2023-JC-QN-0737) (L.F), Natural Science Foundation of Sichuan, China (No. 2023NSFSC1416) (L.F), National Natural Science Foundation of China (Grant No. 62303372) (L.F), Young Talent Fund of Xi'an Association for Science and Technology (No. 959202313033) (L.F), the Fundamental Research Funds for the Central Universities (xzy012024091) (L.F), project funded by China Postdoctoral Science Foundation (No. 2023M742794) (L.F), Postdoctoral Research Project in Shaanxi Province (2023BSHEDZZ34), Shaanxi Province Qin Chuangyuan's "scientist + engineer" program (2024QCY-KXJ-005).

# Appendix A.

## A.1 Experiment Related Images

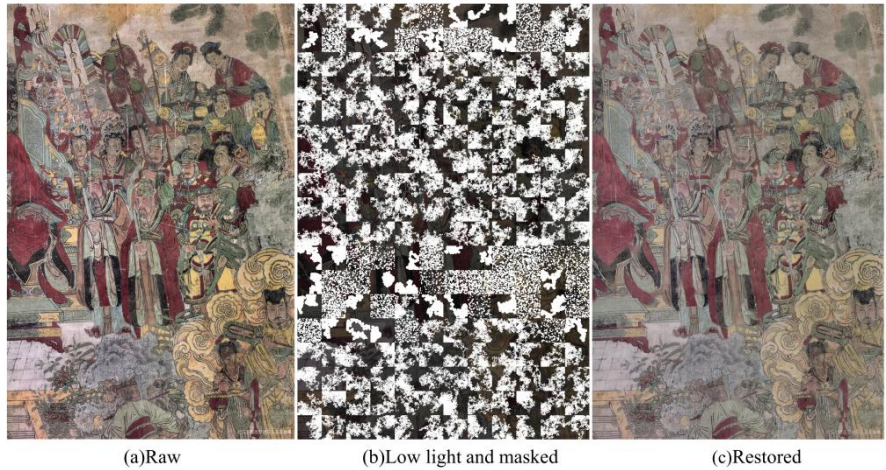

**Fig. S1.** Resulting image after MER network restoration. (a) Raw refers to the original image of an ancient mural painting. (b)Low light and masked refers to the low-light and damaged mural image obtained by reducing the brightness of the original image to 37% and adding random masking to simulate the real archaeological scene. (c) Restored refers to the image after being restored by the MER network.

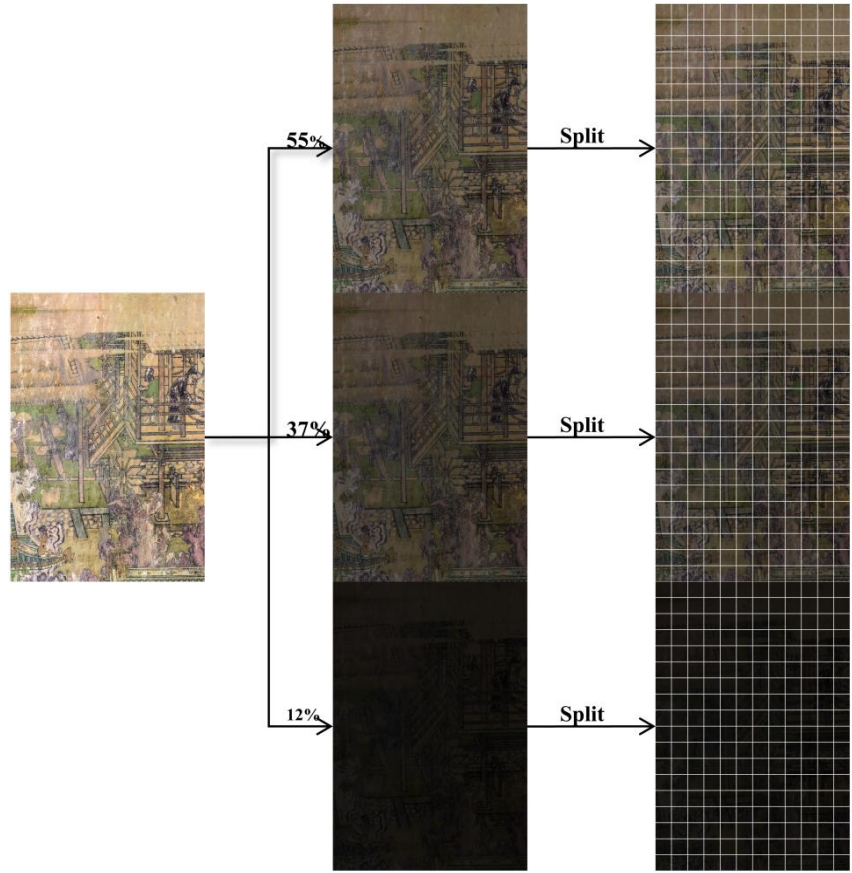

**Fig. S2.** Expansion operations on datasets. Each original mural will be resized to three brightness levels and split into 256×256 pixel sized images.

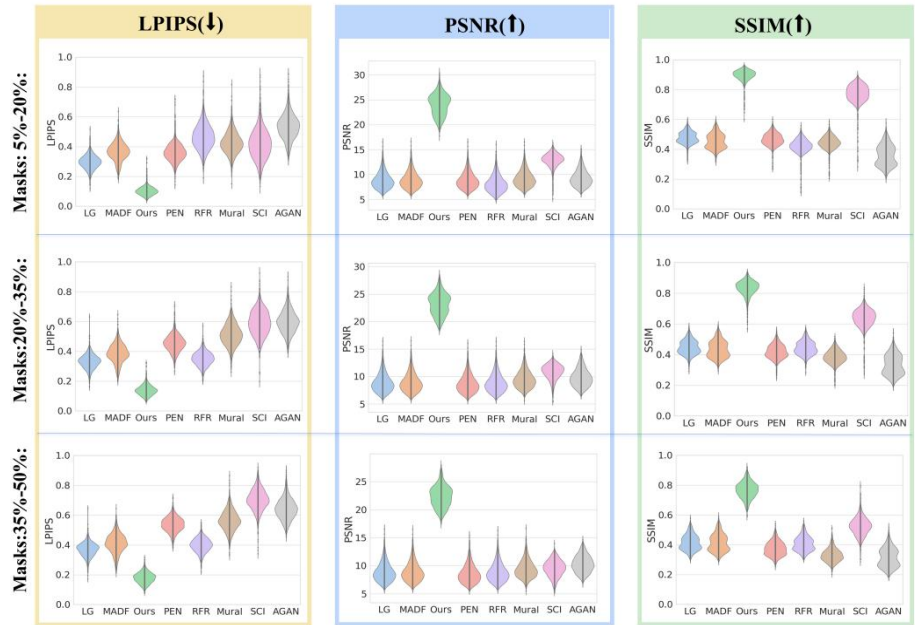

**Fig. S3.** Indicator results of comparative experiments. Boxplots of the metrics evaluating the restorey results of our method with the other six correlation methods for different masking areas when the input luminance is 37% of the ground truth luminance.

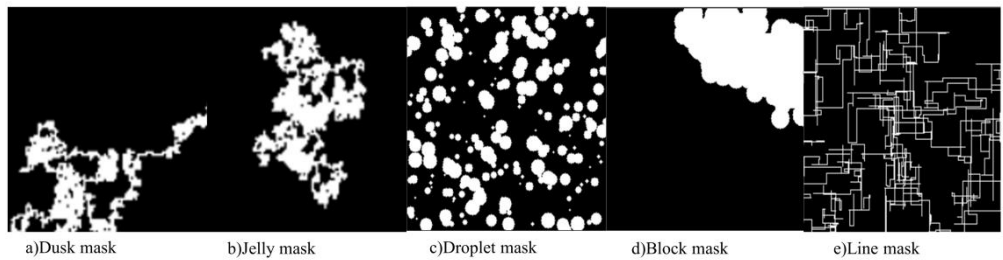

a)Dusk mask  b)Jelly mask  c)Droplet mask  d)Block mask  e)Line mask

**Fig. S4.** Examples of different types of masks. In our experiments we use several defect simulation types to mask the original mural image to simulate the real application scenarios.